\documentclass[conference]{IEEEtran}
\IEEEoverridecommandlockouts
\usepackage{cite}
\usepackage{amsmath,amssymb,amsfonts}
\usepackage{algorithmic}
\usepackage{graphicx}
\usepackage{textcomp}
\usepackage{xcolor}
\usepackage{hyperref,ragged2e}

\def\BibTeX{{\rm B\kern-.05em{\sc i\kern-.025em b}\kern-.08em
    T\kern-.1667em\lower.7ex\hbox{E}\kern-.125emX}}

\begin{document}

\title{NeuroCoreX: An Open-Source FPGA-Based Spiking Neural Network Emulator with On-Chip Learning\\

\thanks{This manuscript has been authored in part by UT-Battelle, LLC under Contract No. DE-AC05-00OR22725 with the U.S. Department of Energy. The United States Government retains and the publisher, by accepting the article for publication, acknowledges that the United States Government retains a non-exclusive, paid-up, irrevocable, world-wide license to publish or reproduce the published form of this manuscript, or allow others to do so, for United States Government purposes. The Department of Energy will provide public access to these results of federally sponsored research in accordance with the DOE Public Access Plan (http://energy.gov/downloads/doe-public-access-plan).This material is based upon work supported by the U.S. Department of Energy, Office of Science, Office of Advanced Scientific Computing Research, under contract number DE-AC05-00OR22725.
}
}

% \author{\IEEEauthorblockN{Ashish Gautam}
% \IEEEauthorblockA{\textit{Computer Science and Mathematics Division} \\
% \textit{Oak Ridge National Laboratory}\\
% Oak Ridge, USA \\
% gautama@ornl.gov}
% \and
% \IEEEauthorblockN{Prasanna Date}
% \IEEEauthorblockA{\textit{Computer Science and Mathematics Division} \\
% \textit{Oak Ridge National Laboratory}\\
% Oak Ridge, USA \\
% datep@ornl.gov}
% \and
% \IEEEauthorblockN{Shruti Kulkarni}
% \IEEEauthorblockA{\textit{Computer Science and Mathematics Division} \\
% \textit{Oak Ridge National Laboratory}\\
% Oak Ridge, USA \\
% kulkarnisr@ornl.gov}
% \and
% \IEEEauthorblockN{Robert Patton}
% \IEEEauthorblockA{\textit{Computer Science and Mathematics Division} \\
% \textit{Oak Ridge National Laboratory}\\
% Oak Ridge, USA \\
% pattonrm@ornl.gov}
% \and
% \IEEEauthorblockN{Thomas Potok}
% \IEEEauthorblockA{\textit{Computer Science and Mathematics Division} \\
% \textit{Oak Ridge National Laboratory}\\
% Oak Ridge, USA \\
% potokte@ornl.gov}
% \and

% }

\author{\IEEEauthorblockN{Ashish Gautam,
Prasanna Date,
Shruti Kulkarni,
Robert Patton,
Thomas Potok}
\IEEEauthorblockA{\textit{Computer Science and Mathematics Division, Oak Ridge National Laboratory},\\ Oak Ridge, Tennessee, USA.\\ {gautama@ornl.gov}}
}

\maketitle

\begin{abstract}
Spiking Neural Networks (SNNs) are computational models inspired by the structure and dynamics of biological neuronal networks. Their event-driven nature enables them to achieve high energy efficiency, particularly when deployed on neuromorphic hardware platforms. Unlike conventional Artificial Neural Networks (ANNs), which primarily rely on layered architectures, SNNs naturally support a wide range of connectivity patterns, from traditional layered structures to small-world graphs characterized by locally dense and globally sparse connections. In this work, we introduce NeuroCoreX, an FPGA-based emulator designed for the flexible co-design and testing of SNNs. NeuroCoreX supports all-to-all connectivity, providing the capability to implement diverse network topologies without architectural restrictions. It features a biologically motivated local learning mechanism based on Spike-Timing-Dependent Plasticity (STDP). The neuron model implemented within NeuroCoreX is the Leaky Integrate-and-Fire (LIF) model, with current-based synapses facilitating spike integration and transmission . A Universal Asynchronous Receiver-Transmitter (UART) interface is provided for programming and configuring the network parameters, including neuron, synapse, and learning rule settings. Users interact with the emulator through a simple Python-based interface, streamlining SNN deployment from model design to hardware execution. NeuroCoreX is released as an open-source framework, aiming to accelerate research and development in energy-efficient, biologically inspired computing.
%NeuroCoreX is an open-source codebase that enables energy-efficient, brain-inspired computing on FPGA hardware. It allows real-time learning, flexible neural connectivity, and general-purpose computing, making neuromorphic technology more accessible for students, researchers, and developers. NeuroCoreX empowers innovation in AI, robotics, healthcare, and smart technology with low-power, adaptable solutions.
\end{abstract}


\section{Introduction}
One of the primary goals of neuromorphic computing is to emulate the structure and dynamics of biological neuronal networks, achieving both brain-like energy efficiency and high computational accuracy. This is accomplished through the use of spiking neuron models implemented on neuromorphic chips. Over the past two decades, a variety of neuromorphic chips have been designed using both analog and digital ASIC platforms, capable of performing real-time information processing \cite{davies2018loihi, mayr2019spinnaker, merolla2014million, benjamin2021neurogrid, miniskar2024neuro, maheshwari2023fpga}. However, the adoption of these systems remains constrained by their high cost, limited availability, and architectural specificity. Proprietary neuromorphic chips typically restrict user access and customization, creating significant barriers for researchers and students seeking to innovate and explore new designs.

Field programmable gate arrays (FPGAs) offer a promising alternative, providing a flexible platform for prototyping and validating SNNs before final implementation on custom ASICs. They serve as an effective intermediate step, facilitating co-design development alongside off-the-shelf SNN simulators \cite{date2023superneuro, stimberg2019brian}. Several digital ASICs and FPGA-based SNN systems have been proposed in the past \cite{mitchell2020caspian, liu2022fpga, matinizadeh2024neuromorphic}. While some proprietary systems \cite{davies2018loihi} include local learning capabilities such as spike-timing-dependent plasticity (STDP), most FPGA-based implementations still rely heavily on offline training and lack real-time, on-chip learning. This limitation reduces their adaptability for dynamic, continuously evolving applications such as robotics, smart sensors, and edge computing.

To address these challenges, we introduce NeuroCoreX, an open-source spiking neural network (SNN) emulator implemented in VHDL (Very High-Speed Integrated Circuit Hardware Description Language) for FPGA platforms. NeuroCoreX provides an affordable and flexible alternative for neuromorphic computing research and education. 
It is meant to be used in AI applications requiring low size, weight, and power (SWaP) such as edge computing, embedded systems, Internet of Things (IoT), and autonomous systems \cite{patton2022neuromorphic,kulkarni2023sensor,aimone2022review,cong2022semi,cong2023hyperparameter}.
Unlike fixed-architecture hardware, it supports fully reconfigurable network topologies, from simple layered structures to complex small-world graphs. It incorporates biologically inspired local learning through a variant of the STDP learning rule \cite{markram1997regulation}, enabling on-chip, online adaptation of synaptic weights. The system uses a Leaky Integrate-and-Fire (LIF) neuron model with current-based synapses \cite{gerstner1995time}, ensuring both computational simplicity and biological relevance.
This model of neuromorphic computation is known to be Turing-complete, i.e., capable of performing all the computations that a CPU/GPU can perform \cite{date2022neuromorphic, date2021computational}.
As a result, NeuroCoreX can support not just SNN-based AI workloads but also general-purpose computing workloads \cite{date2023encoding,date2022virtual,schuman2021sparse,wurm2023arithmetic}.

Programming and configuring NeuroCoreX is streamlined through a UART interface and a simple Python module, allowing users to modify network, neuron, synapse, and learning parameters easily. This makes NeuroCoreX not only a valuable research tool for testing new theories of learning and network organization but also a powerful educational platform for hands-on experience with neuromorphic hardware. Additionally, its energy-efficient architecture makes it well-suited for low-power AI applications in areas such as autonomous systems, smart sensors, and scientific instrumentation.

%The remainder of this paper is organized as follows: Section II describes the hardware architecture of NeuroCoreX in detail. Section III discusses the implementation of local learning mechanisms. Section IV presents case studies and performance evaluations. Finally, Section V concludes the paper and outlines future directions.
The rest of the manuscript is organized as follows: Section \ref{sec:neurocorex} provides an overview and the architecture description of NeuroCoreX in detail. In Section \ref{sec:results}, we present the results demonstrating the functionality of the platform and evaluate its performance on the DIGITS dataset~\cite{optical_recognition_of_handwritten_digits_80}. The manuscript concludes with a discussion of the results and planned future work in Section \ref{sec:conclusion}.

\section{NeuroCoreX}
\label{sec:neurocorex}

\subsection{NeuroCoreX overview}
NeuroCoreX is designed to emulate brain-like computation on reconfigurable FPGA hardware using a digital circuit approach. The system architecture is built around three fundamental components, inspired by biological neural networks: neurons, synapses, and a local learning mechanism. These elements are digitally realized in VHDL and operate together to support real-time, adaptive information processing.

The neuron model employed is the LIF model, which captures the essential dynamics of biological neurons with computational efficiency and is known to be Turing-complete. Synapses are modeled with an exponential current response and store dynamic weight values that govern neuron-to-neuron influence. Learning is enabled through a simple variant of STDP, allowing synaptic strengths to evolve based on the relative timing of neuronal spikes.

In its current implementation, NeuroCoreX supports networks of up to $N=100$ neurons with full all-to-all bidirectional connectivity using 10,000 synapses. In addition to recurrent connections, the system includes a separate set of 10,000 feedforward input synapses that serve as the interface for external stimuli. These input weights determine how incoming spikes—from sources such as sensors or preprocessed datasets—modulate the activity of neurons within the network. Neuronal dynamics are configured to emulate biological timescales. The network size and acceleration factor can be scaled depending on the memory resources of the FPGA, precision of the synaptic weights used and the operating clock frequency. Time-multiplexing and pipelining techniques are used to optimize hardware resource usage. A single physical neuron circuit is time-multiplexed to emulate the entire network. Communication with the FPGA is managed through a UART interface, with a Python module providing a user-friendly configuration and control interface.
\begin{figure}[t]
  \centering
  \includegraphics[width=\linewidth]{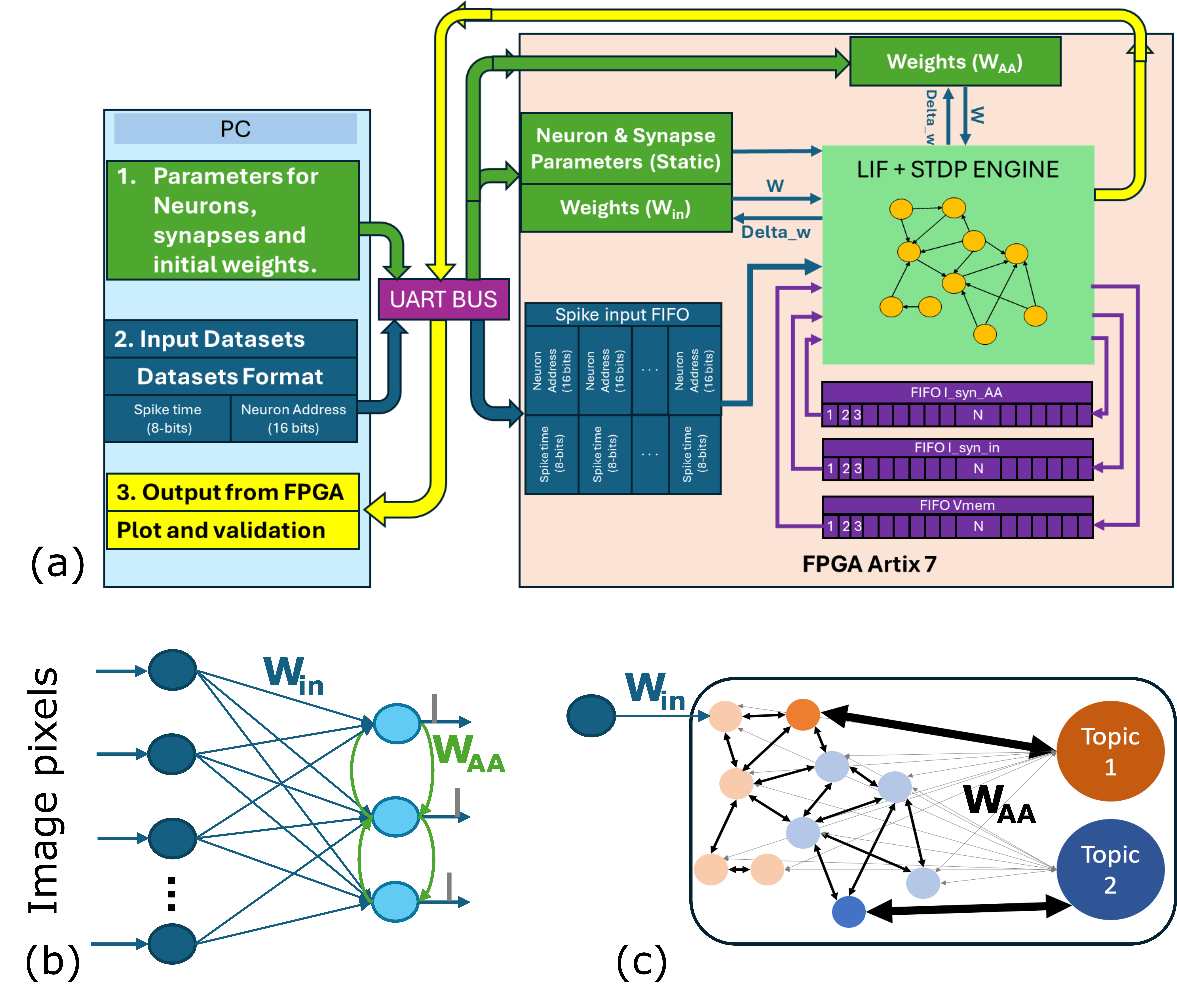}
  % \includegraphics[width=\linewidth]{images/NeuroCoreX.png}
  % \includegraphics[width=\textwidth]{images/Diagram.png}
  % \caption{Block diagram}
  \caption{(a). Block diagram of our FPGA based NeuroCoreX, (b). Feedforward SNN used for digits dataset classification, (c). Spiking Graph Neural Network for citation graph node classification problem.}

  \label{fig:fig1}
\end{figure}
The operation of NeuroCoreX follows a structured emulation cycle (see Fig.~\ref{fig:fig1}(a)). First, the network weights and initial configuration parameters for neuron, synapse, and learning rule are transferred from a PC to the FPGA via the UART interface. Once the network is set up, input spikes are streamed in real time, buffered in a First-In-First-Out (FIFO) module on the FPGA, and injected into the network precisely at their intended timestamps. At each neuron update cycle, the time-multiplexed processor sequentially updates the membrane potential, synaptic inputs, and firing status of each neuron. If a neuron fires, its effect on connected neurons is mediated through the all-to-all connected $W_{AA}$  weight matrix, and synaptic weights are updated in real time if STDP is enabled. Synaptic weights corresponding to feedforward inputs $W_{in}$ are similarly updated if STDP is enabled for them. The system thus continuously processes incoming spikes, updates network states, applies learning, and advances to the next time step, enabling real-time emulation of SNNs on the FPGA. %Figure~\ref{fig:fig1} shows a high level block diagram of NeuroCoreX. We implement the trained network for digits dataset ~\cite{optical_recognition_of_handwritten_digits_80}. for inference on the FPGA, which has feedforward topology. The citation graph node classification SNN has a non-layered topology, and the $W_{AA}$ weight matrix is used to realize the weights both plastic and static ones.

Fig.~\ref{fig:fig1}(a) shows a high-level block diagram of NeuroCoreX, along with two representative examples of network architectures that can be implemented on the FPGA. The first is a conventional feedforward SNN, a topology commonly used in neuromorphic research. We use this network to demonstrate digit classification on the well-known DIGITS dataset~\cite{optical_recognition_of_handwritten_digits_80}, showcasing NeuroCoreX’s support for standard inference tasks. The second network, shown in Fig.~\ref{fig:fig1}(c), illustrates a SNN designed for node classification on citation graphs using STDP-based unsupervised learning. This architecture lacks a traditional layered structure and is instead defined by arbitrary, sparse connectivity encoded in the 
$W_{AA}$ matrix, which stores both plastic and static synaptic weights.

These two examples highlight the flexibility of NeuroCoreX: in addition to supporting conventional layered architectures, the platform can implement non-layered networks such as those found in graph-based problems or generated via evolutionary algorithms like EONs \cite{}. This versatility makes it suitable for a wide range of neuromorphic applications, from structured inference tasks to irregular and adaptive network topologies.
%Together, these features make NeuroCoreX a versatile, low-power platform for neuromorphic research, education, and edge AI applications.

\subsection{FPGA platform}
NeuroCoreX is implemented on the Artix-7 FPGA, a cost-effective and widely available platform that offers sufficient resources for neuromorphic prototyping. The system operates with a maximum internal clock frequency of 100 MHz. Two main clock domains are used: the high-frequency 100 MHz clock for UART-based communication and a 100 KHz lower-speed operating clock for neural processing. The combination of modest resource requirements, real-time adaptability, and biological plausibility makes the Artix-7 platform an ideal choice for NeuroCoreX. Scalability to larger networks or faster processing rates is primarily limited by the available block RAM and choice of clock frequency for neural processing on the FPGA.

The UART interface operates at a baud rate of 1 Mbps, enabling efficient transmission and reception of both static configuration data and real-time input spikes. The FPGA receives network weights, neuron, synapse, and learning parameters from a host PC via this UART channel before execution begins. During operation, additional input spikes are streamed to the network in real time through the same interface.

%To maximize hardware efficiency, the neuron array is implemented using time-multiplexing and pipelining techniques. Instead of dedicating separate hardware resources for each neuron, a single neuron processing unit sequentially cycles through the entire network state, updating each neuron and its associated synaptic interactions within a pipelined structure.

\subsection{Neuron and  Synapse Models}
NeuroCoreX employs a biologically inspired computational model that integrates neuron dynamics, synaptic interactions, and local learning mechanisms. 
%All parameters associated with neurons, synapses, and learning rules can be programmed via the UART interface, enabling flexible experimentation. 
The neurons are modeled using a LIF formulation, adapted for efficient FPGA implementation. Each neuron has four configurable parameters, threshold, leak, refractory period, and reset voltage. The membrane potential V(t) is updated at each time step according to the following discrete-time equation:
\[
V(t+1) = V(t) - \lambda + I_{\text{syn}}(t)
\]

where $I_{\text{syn}}(t)$ is the total synaptic input current at timestep $t$ and $\lambda$ is the neuron's leak. When the membrane potential exceeds the threshold $V_{\text{th}}$, the neuron emits a spike, enters a refractory period $\tau_{\text{ref}}$, and its membrane potential is reset to $V_{\text{reset}}$. To ensure efficient real-time processing, all calculations are performed using a fixed-point format with 1 sign bit, 7 integer bits, and 10 fractional bits.

Synaptic inputs are modeled as current-based exponential synapses, capturing biologically realistic, temporally decaying post-synaptic responses. The synaptic current dynamics follow the update rule:
\[
I_{\text{syn}}(t+1) = I_{\text{syn}}(t) - \lambda_{\text{syn}}
\]

where $\lambda_{\text{syn}}$ represents the synaptic current decay at each time step. Each synapse has an associated weight that determines its influence on the postsynaptic neuron. These synaptic weights, stored in BRAM, are dynamically updated during runtime. Weights are represented in signed 8-bit format, and appropriate resizing and bit-shifting are applied during computations to correctly integrate the synaptic current into the membrane potential.

\begin{figure}[h]
  \centering
  \includegraphics[height=0.38\textheight]{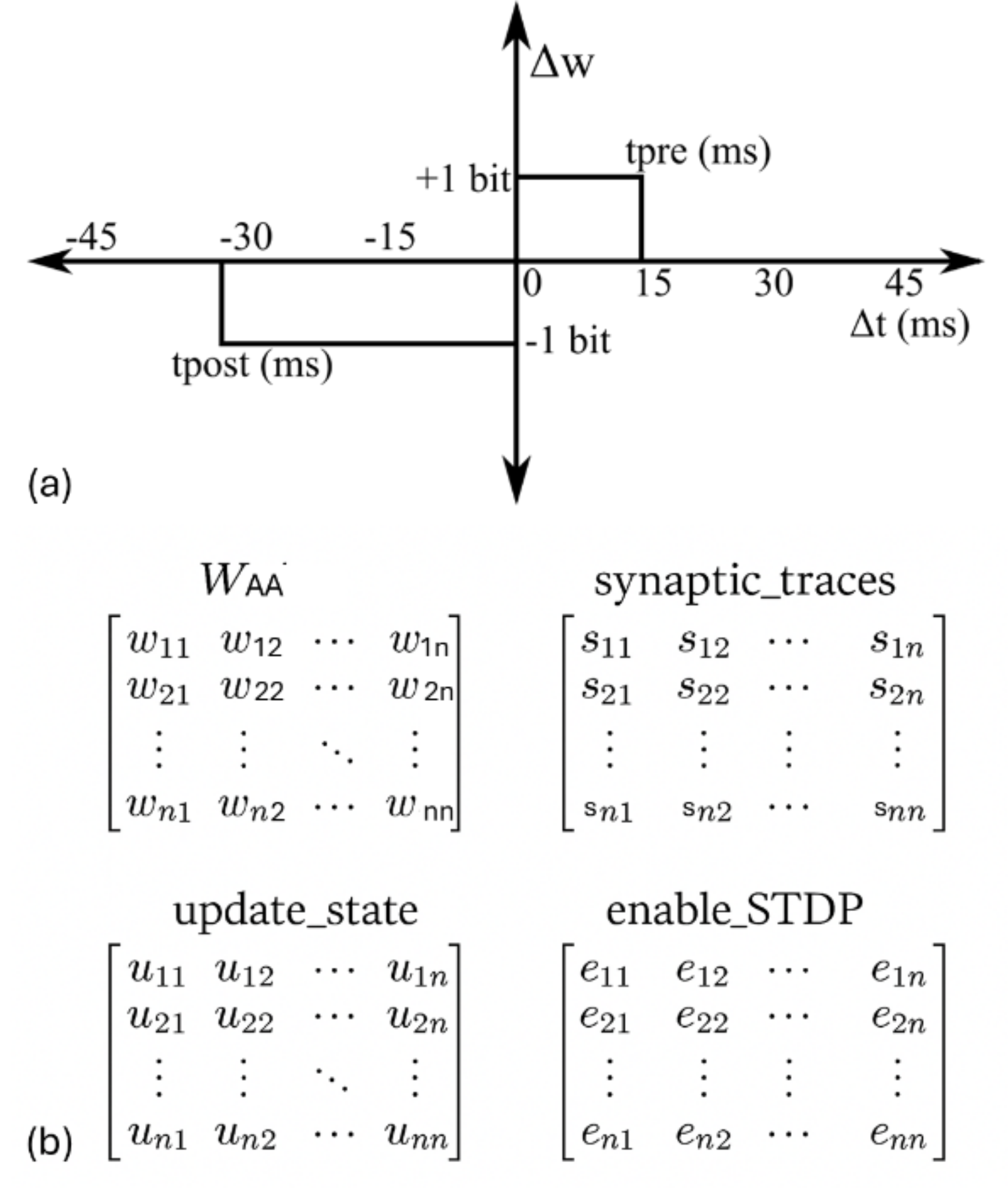}
  \caption{(a) Simplified STDP learning rule implemented on NeuroCoreX. (b) Internal variables stored in BRAM for tracking STDP updates. All matrices are of size \( N \times N \) and stored in row-major order. Each element of the \( W_{AA} \) and \texttt{synaptic\_traces} matrices is 8 bits wide, while \texttt{update\_state} and \texttt{enable\_STDP} are binary matrices.}

  \label{fig:fig2}
\end{figure}

\subsection{Learning Rule}
NeuroCoreX implements a pair-based STDP rule using a rectangular learning window, a simplification widely demonstrated to maintain similar functional behavior to conventional exponential window \cite{markram1997regulation} when the resolution of the weights is greater than 6-bits \cite{gautam2023adaptive,gautam2021adaptive,cassidy2011combinational}. In this model (See Fig.~\ref{fig:fig2} (a)), if a causal spike pair (presynaptic spike preceding postsynaptic spike) occurs within the window $t_{pre}$, the synaptic weight is incremented by $dw_{pos}$. If an acausal spike pair (postsynaptic spike preceding presynaptic spike) occurs, the weight is decremented by $dw_{neg}$.

\[
\Delta w =
\begin{cases}
dw_{\text{pos}}, & \text{if} \quad 0 <  \Delta t < t_{\text{pre}} \\
- dw_{\text{neg}}, & \text{if} \quad 0 <  \Delta t < t_{\text{post}} \\
0, & \text{otherwise}
\end{cases}
\]
Here, $\Delta t$ is the time difference between pre-and postsynaptic spikes, \( dw_{\text{neg}} \), \( dw_{\text{pos}} \), \( t_{\text{pre}} \), and \( t_{\text{post}} \) are configurable parameters initialized via the Python-UART interface. The pre- and post-synaptic spike timings are tracked using dedicated time-trace registers stored in BRAM (see Fig.~\ref{fig:fig2}(b)). These time traces are updated on each spike event and are used to detect causal or acausal spike pairings that trigger weight updates.

For example, when neuron 1 spikes, all synaptic traces corresponding to its outgoing synapses are reset to zero, and the associated \texttt{update\_state} entries are set to 1. In parallel, the post-synaptic trace (not shown in Fig.~\ref{fig:fig2}) is activated for all incoming synapses to neuron 1. At each subsequent time step, the active values in the synaptic trace matrices are incremented by 1. This process continues until the counter reaches \( t_{\text{pre}} \). If no other neuron spikes within this window, the trace value is reset to \texttt{0xFF} (representing a negative disabled state), and the corresponding \texttt{update\_state} entry is cleared. Similarly, if no neuron spiked within \( t_{\text{post}} \) time steps prior to neuron 1's spike, the post-synaptic trace is also reset to a negative value.

However, if another neuron spikes within \( t_{\text{pre}} \) time steps after neuron 1, the synaptic weight is incremented by \( dw_{\text{pos}} \), and both the synaptic trace and \texttt{update\_state} for that synapse are reset. Conversely, if a neuron spiked within \( t_{\text{post}} \) time steps prior to neuron 1, the synaptic weight is decremented by \( dw_{\text{neg}} \), and the associated trace and \texttt{update\_state} values are reset. Thus, during each neuron update cycle, if a neuron spikes, the corresponding row and column addresses in the matrices shown in Fig.~\ref{fig:fig2}(b) are accessed and updated. Based on the current states of these auxiliary matrices, the entries in the weight matrix \( W_{AA} \) are modified accordingly.

The \texttt{enable\_STDP} matrix is a static binary mask configured via the Python interface at initialization. It acts as a filter to specify which synapses in \( W_{AA} \) are subject to STDP-based plasticity. There a similar matrix for synapses in \( W_{in} \).

\subsection{Network Architecture and System Operation}
The SNN architecture implemented on NeuroCoreX is illustrated in Fig. ~\ref{fig:fig1}(a). The network consists of upto $N=100$ LIF neurons instantiated on the FPGA. Two primary weight matrices define the network connectivity: $W_{AA}$, a synaptic weight matrix for all-to-all, bidirectional connectivity between neurons in the FPGA and $W_{in}$, a synaptic weight matrix for feedforward connections from external input sources to the neurons on the FPGA. Both matrices are stored in the FPGA’s BRAM. They can be initialized to user-defined values, and are accessed during SNN emulation. A synaptic weight value of zero indicates no connection between the corresponding neurons. Internal network dynamics are governed by the $W_{AA}$ matrix. This matrix allows every neuron to influence every other neuron bidirectionally. The matrix values determine the synaptic strengths between pairs of neurons and evolve over time via STDP-based learning. Both $W_{AA}$ and $W_{in}$ matrices support on-chip learning. To preserve network structure and prevent unwanted modifications, an associated binary filter matrix, called enable-STDP, is used for each weight matrix. If a weight’s corresponding enable-STDP entry is zero, the weight remains fixed throughout operation—even during learning phases. Weights representing nonexistent connections (zeros in the weight matrix) are thus protected from modification.
In addition to the synaptic weights, the BRAMs are also used store the pre-and post synaptic traces necessary for STDP calculations. Weight matrices $W_{AA}$ and $W_{FF}$, and the pre-and postsynaptic traces are stored as separate memory banks in row major order on the FPGA's BRAM. As BRAM addresses must be accessed sequentially, the high-speed 100 MHz clock domain is utilized for reading, updating, and writing synaptic weights. During each clock cycle, synaptic weights and neuron states are updated in a pipelined manner to ensure efficient processing without data bottlenecks.

NeuroCoreX utilizes time-multiplexing and pipelining techniques to emulate 100 neurons using a single physical neuron processing unit. Neuron updates are managed under a 100 kHz clock domain, such that updating all 100 neurons takes 1 millisecond, which closely matches the biological timescale of real neural systems. To accelerate the emulation, a higher update clock frequency can be used. For example, operating the neuron updates at 1 MHz results in a 10× speed-up relative to biological time for a network of 100 neurons. However, if the network size is increased to 1000 neurons while maintaining the 1 MHz clock, the full network would again require approximately 1 millisecond per time step, restoring biological equivalence. Thus, there exists a direct dependence between the number of neurons, the update clock frequency, and the effective emulated timescale. In the current implementation, the network size is limited to 100 neurons due to the available BRAM resources on the Artix-7 FPGA. Even with higher BRAM availability, the number of neurons that can be emulated is ultimately constrained by the difference between the clock frequency available for BRAM access and the clock rate used for updating the SNN states (see Section \ref{sec:conclusion}).

Incoming spikes from external sources are transmitted via the UART interface.  Each spike is encoded as a 24-bit word, comprising a 16-bit input neuron (or pixel) address and an 8-bit spike timing component. In the current implementation, the feedforward weight matrix \( W_{\text{in}} \) is a \( 100 \times 100 \) matrix, corresponding to 100 input neurons and 100 on-chip neurons. Although 8 bits are sufficient to encode the addresses of 100 input neurons, we chose to use 16 bits to provide flexibility for interfacing with larger sensor arrays. This allows the system to support up to 16K input neurons in future applications. In such cases, the feedforward matrix \( W_{\text{in}} \) becomes a rectangular matrix of size \( 100 \times N_{\text{in}} \), where \( N_{\text{in}} \) denotes the number of input neurons in the external layer.
For transmission efficiency, successive time differences between spikes are sent, rather than absolute times. These incoming spikes are temporarily stored in a FIFO buffer on the FPGA (See Fig.~\ref{fig:fig1}(a)).  The FIFO is designed to support simultaneous read and write operations, allowing it to continuously receive long temporal spike trains while concurrently feeding data to the network in real time without stalling. During network emulation, the system clock continuously increments an internal time counter. When the internal clock matches the timestamp of the spike at the FIFO head, the corresponding input neuron address is read. The associated weights from the $W_{in}$ matrix are then used to inject synaptic currents into the membrane potentials of the relevant neurons. If the synaptic current causes any neuron on the FPGA to spike, then associated weights from the $W_{AA}$ matrix are then read and used to inject synaptic currents into the membrane potentials of all other neurons connected to the spiking neuron in the network.

\section{Results}
\label{sec:results}

We present experimental results that demonstrate the usability, correctness, and flexibility of the NeuroCoreX platform across a range of SNN workloads.

% We now present the results from realizing the trained networks for image classification task. %and node classiicaton tasks for citation graphs.
% %\subsubsection{Microseer}
% For the classification task, we use the digits dataset from UCI repository~\cite{optical_recognition_of_handwritten_digits_80} through the scikit-learn library. The following subsections describe the training on software and inference results on NeuroCoreX.

\subsection{Demonstrating User Interface Flexibility}

One of the key strengths of the NeuroCoreX platform lies in its flexible and intuitive user interface, which enables seamless communication between a host PC and the FPGA hardware through a Python-based control module. To demonstrate this capability, we highlight several core features supported by the interface.

\begin{figure}[h]
  \centering
  \includegraphics[width=\linewidth]{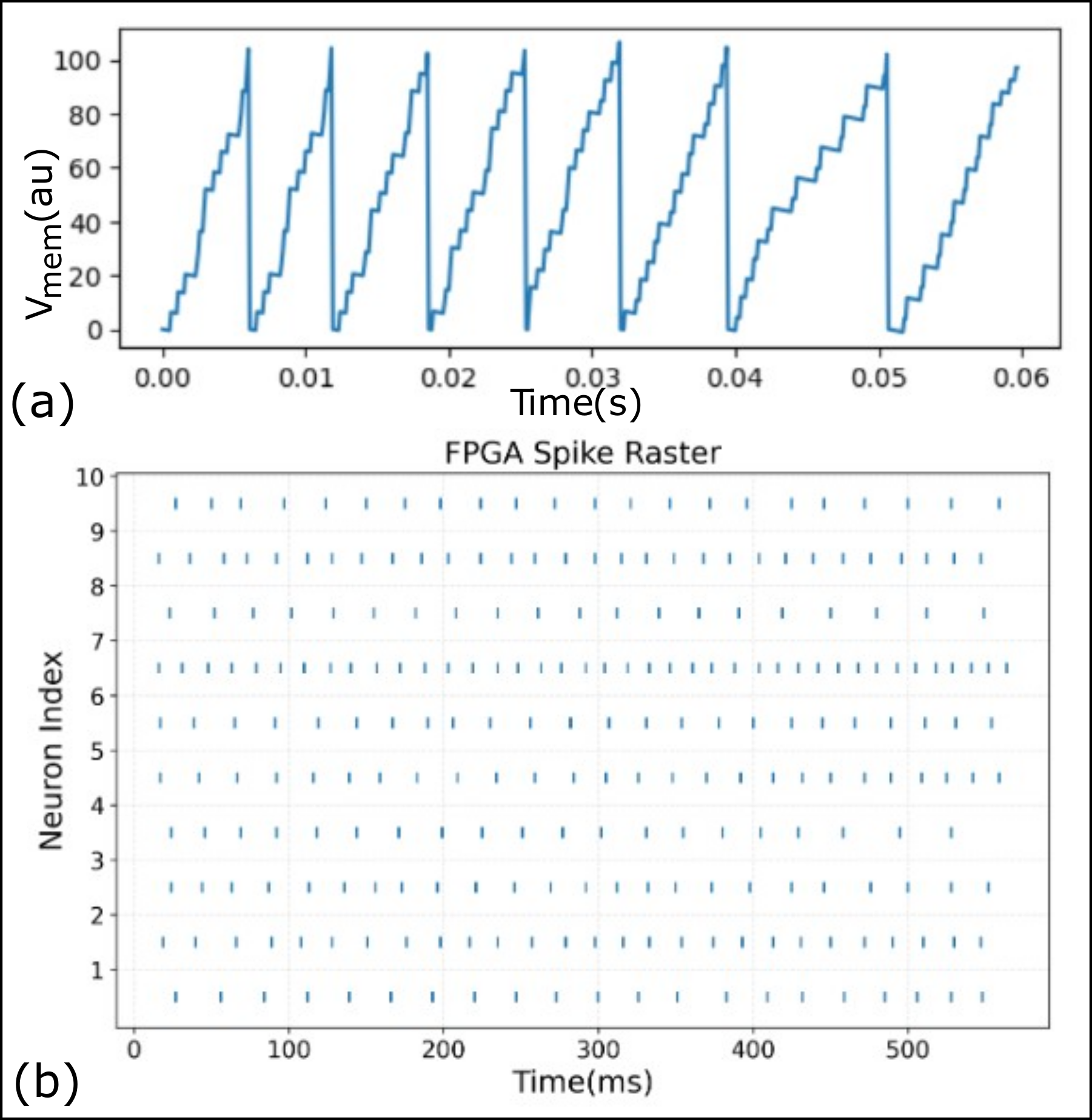}
  \caption{(a) Membrane potential trace of a selected neuron, recorded from the FPGA during network emulation. (b) Spike raster plot showing activity of 10 neurons during a test run. Both plots were generated using data read back from the FPGA, demonstrating the observability and debugging capabilities of the NeuroCoreX interface.}

  \label{fig:fig3}
\end{figure}

First, spike trains can be streamed from the PC to the FPGA for real-time emulation of spiking network activity. Figure~\ref{fig:fig3}(b) shows a raster plot of spiking activity recorded during one such emulation run. Second, the user interface allows for real-time monitoring of internal neuron dynamics. Specifically, the membrane potential of any selected neuron can be captured and plotted as a time series, offering insight into subthreshold integration and spiking behavior. Figure~\ref{fig:fig3}(a) shows the membrane potential trace of a representative neuron under input stimulation. The spike trains of all neurons and the membrane potential of a selected neuron are transferred back to the PC in real time. These signals are not stored on the FPGA. Instead, an internal FIFO module is used to buffer these signals, which allows for the continuous recording and visualization of network dynamics over long temporal duration without being limited by on-chip memory. Finally, the interface supports reading back synaptic weights from the FPGA after an STDP-enabled emulation. This feature enables direct inspection of weight evolution and verification of learning dynamics on hardware. It is particularly useful for comparing hardware learning outcomes with software simulations, facilitating debugging and model validation.

%These capabilities enable a tight co-design loop between SNN model development, hardware validation, and analysis—making NeuroCoreX not only a neuromorphic inference engine but also a highly configurable research tool.

%First, spike trains can be streamed from the PC to the FPGA for emulating network activity in real time. Figure~\ref{fig:fig3}(b) shows an example raster plot of neuron activity generated after running a test spike train on the FPGA while activating 10 input neurons. Second, the interface allows users to monitor the internal dynamics of individual neurons during emulation. Specifically, membrane potentials can be recorded from a selected neuron and plotted for visualization. Figure~\ref{fig:fig3}(a) shows the membrane potential trace of a representative neuron under input stimulation. The spike trains of all neurons and membrane potential of one chosen neuron are transferred in real time and are not stored on the FPGA, thus raster plot of very long temporal scales can be recorded from the network. Third, the interface provides support for reading back updated synaptic weights, allowing verification of learning dynamics during STDP-enabled runs. This feature is particularly useful for debugging, analyzing weight evolution, and validating hardware learning behavior against expected outcomes from simulation. 
These features collectively support efficient testing, inspection, and refinement of neuromorphic models, enabling a co-design loop between high-level model development and hardware validation.

\subsection{DIGITS Dataset}

% \subsection{Inference on the DIGITS Dataset}

To verify the functional correctness of internal neuron and synapse computations on the FPGA, we performed inference on the DIGITS dataset ~\cite{optical_recognition_of_handwritten_digits_80} using a model trained in the SuperNeuroMAT simulator~\cite{date2023superneuro}. The dataset contains a total of 1,797 samples of handwritten digits, each represented as an \( 8 \times 8 \) grayscale image with pixel values in the range \([0, 15]\). The dataset was split into 70\% training samples and 30\% test samples.

\begin{figure}
    \centering
    \includegraphics[width=0.4\linewidth]{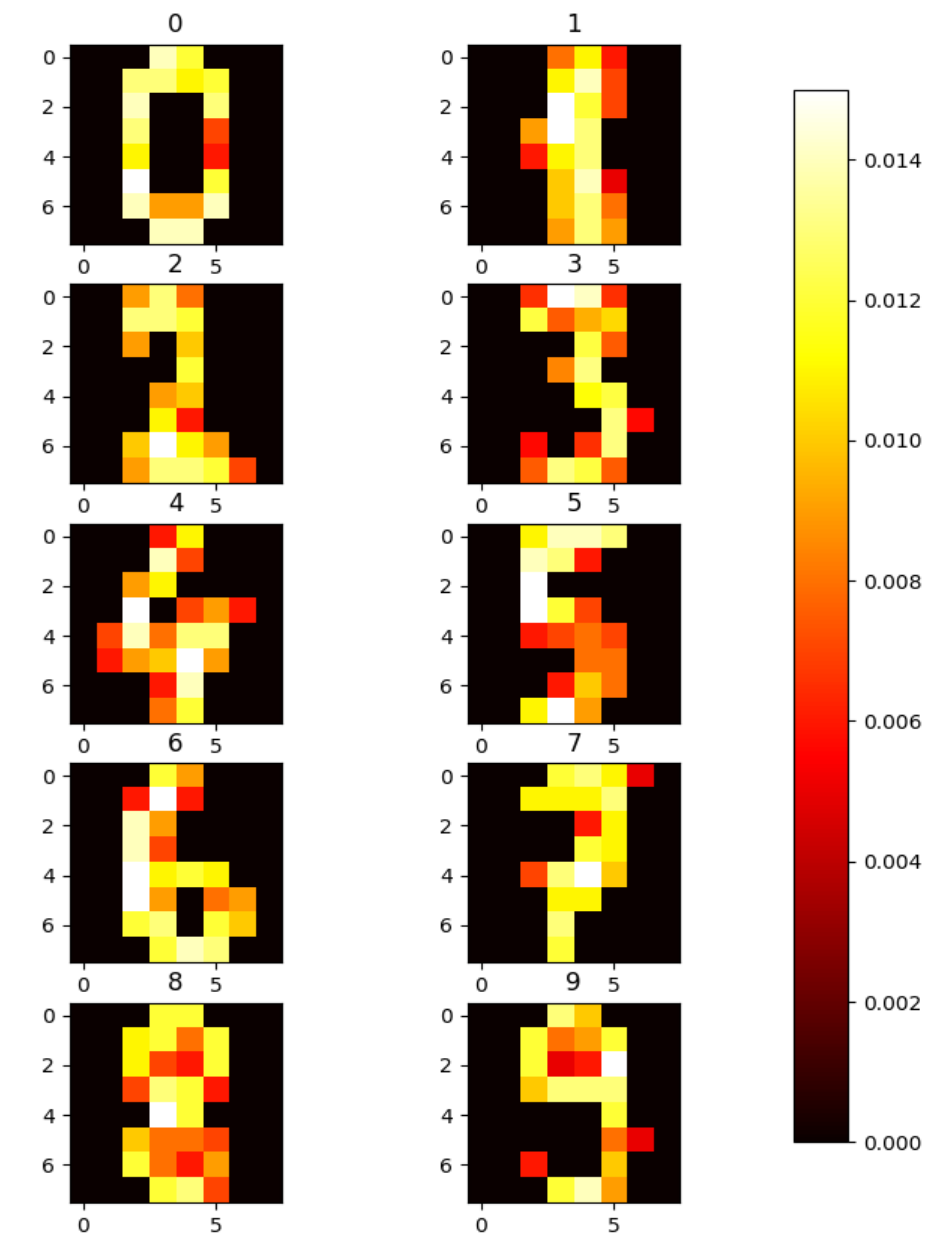}
    \caption{Heat-map of the trained weights from all the $10$ output neurons.}
    \label{fig:trained_weights}
\end{figure}

A simple two-layer feedforward spiking neural network was trained using integrate-and-fire neurons and weighted synapses in the SuperNeuroMAT simulator. The input images were normalized to the range \([0, 1]\) and converted into input spike trains using rate encoding. Each pixel intensity was encoded as twice the number of spikes and distributed uniformly over 32 time steps. During training, target labels were encoded by delivering a spike to the corresponding output neuron at timestep \( t + 1 \), one timestep after the input presentation at \( t \). The learning was carried out using one-shot STDP-based training. It is important to note that the training was not aimed at maximizing classification accuracy; rather, the goal was to validate the correctness of the internal neuron and synapse dynamics on the FPGA platform. Figure~\ref{fig:trained_weights} shows the final weights of the output neurons after training, which clearly reflect the digit-specific patterns learned by the network.

For deployment on NeuroCoreX, the trained weights and network parameters were transferred and initialized on the FPGA. The STDP learning rule was disabled during this phase to maintain fixed weights. The identical spike sequences for the test set were streamed to the FPGA through the UART interface. %Each test sample was presented for 32 time steps using the same rate encoding as in simulation.
We achieved a test accuracy of 68\% on the SuperNeuroMAT simulator, and the same accuracy was observed on the NeuroCoreX hardware. This result confirms that the FPGA implementation faithfully reproduces the dynamics of the simulated SNN and validates the correctness of internal spike integration, thresholding, and synaptic current accumulation in hardware.

%\subsubsection{Training on SuperNeuroMAT} 

%The digits dataset has a total of $1,797$ samples, which were split as $70\%$ training samples and $30\%$ test samples. Each sample is an image of size $8\times8$, with pixel values in the range $[0,15]$. We train a simple two-layered feedforward SNN on the SuperNeuroMAT simulator \cite{date2023superneuro}, using integrate and fire neurons and weighted synapses. 
%During training, the image pixels were normalized to $[0,1]$ and applied to the first layer of neurons as weighted input spikes at timestep $t$. The targets are set by spiking the label neurons in the output layer with a spike in the next timestep $t+1$. We use a one-shot learning to train the weights of the network using the in-built STDP learning rule. Figure~\ref{fig:trained_weights} show the weights of the output neurons after training. 
%These images correctly depict the digits from 0--9 and validate that STDP learning has occurred correctly on SuperNeuroMAT.
% \begin{figure}[t]
%     \centering
%     \includegraphics[height=0.17\textheight]{images/trained_wts_hist_inference.pdf}
%     \caption{Heat-map of the trained weights from all the $10$ output neurons.}
%     \label{fig:trained_weights}
% \end{figure}

%\subsubsection{Inference on NeuroCoreX} 
%We further scale up the weights and apply rounding to get them in the integer value. Figure~\ref{fig:w_hist} shows the histogram of weights used for running inference on the NeuroCoreX. 
%For running the inference, the STDP was turned off in the network. The inputs were encoded into spikes  using rate encoding. Each sample was presented for 32 timesteps, with each pixel being encoded into two times the number of spikes as the corresponding pixel intensity. 
% \begin{equation}
%     S_i(t) = [1 for i in range(x_i)]
% \end{equation}
%We achieved an accuracy of $68\%$ on the digits test set of $594$ samples on SuperNeuroMAT software simulation. The same network configuration was programmed on NeuroCoreX through the UART interface, and the same set of spikes corresponding to the test set were transmitted to the FPGA. The inference run on NeuroCoreX also gave us the exact same test accuracy of $68\%$. Note that in this work we focused on validating the equivalency of the network simulated in software and on the hardware. 

% \begin{figure}
%     \centering
%     \includegraphics[width=0.65\linewidth]{images/infernce_weights_histogram.png}
%     \caption{Histogram of trained weights after scaling and rounding. These weights are used for inference.}
%     \label{fig:w_hist}
% \end{figure}

% \begin{itemize}
%     \item Dataset from scikit-learn
%     \item Training with STDP as one shot learning
%     \item figures for trained network weights
%     \item Data encoding for inference
%     \item 
% \end{itemize}

\subsection{MicroSeer Dataset}

% \textcolor{red}{Write a brief description of Citation graph and microseer dataset and the way evaluation was performed on SNMAT}

To evaluate the applicability of NeuroCoreX for graph-based learning tasks, we tested its performance using the MicroSeer dataset. MicroSeer is a reduced version of the Citeseer citation graph~\cite{caragea2014citeseer}, containing 84 papers labeled with six topic categories. It was constructed by iteratively removing nodes from the largest connected component of Citeseer while ensuring that the resulting graph remained a single connected component. This connectivity was prioritized because it is assumed that learning from a very small, fragmented dataset would be ineffective. This reduction process yielded a total of 90 neurons, making the dataset well suited for deployment on NeuroCoreX, which supports up to 100 neurons and 10,000 bidirectional synapses.

%The MicroSeer dataset is a significantly reduced dataset derived from the largest connected component of the Citeseer dataset \cite{caragea2014citeseer}. 
%It contains 84 papers and their six associated topics. 
%This results in a total of 90 neurons in the SNN, making it small enough to be accommodated on NeuroCoreX. 
%with limited neuron capacity. 
%MicroSeer was created by randomly deleting papers from the largest connected component of Citeseer while ensuring the remaining network stayed connected as a single component. 
%This connectivity was prioritized because it's assumed that learning from a very small, fragmented dataset would be ineffective. 

%Unlike traditional supervised learning approaches that rely on iterative gradient-based optimization, our method constructs the SNN directly from the graph structure and trains it using spike-timing-dependent plasticity (STDP). During classification, a spike is injected into the neuron representing the test paper. This initiates a cascade of spikes through the graph, during which STDP modifies the synaptic weights between paper and topic neurons. The final classification is determined by identifying the topic neuron with the highest post-learning synaptic strength to the test paper neuron.

As compared to standard supervised learning that uses iterative error correction for weight updates, our training method leverages the graph's structure directly to build the network. 
When testing a paper in the test data set, spiking the neuron associated with the test paper triggers a chain reaction of spikes. 
As these spikes travel between paper and topic neurons, STDP dynamically modifies the weights of the synapses connecting the test paper neuron to the topic neurons and vice versa. 
Subsequently, classification is achieved by finding the topic neuron with the highest final synaptic strength from the test paper neuron under consideration. 
The topic corresponding to this topic neuron is the one predicted by the SNN for the given test paper.

%The model was first trained and validated in the SuperNeuroMAT simulator and then ported to the NeuroCoreX hardware. When STDP was disabled, inference results on the FPGA matched those from the simulator, confirming functional equivalence. However, with STDP enabled, the weight updates diverged due to differences in implementation: SuperNeuroMAT uses an exponential STDP rule with 64-bit floating-point weights, whereas NeuroCoreX employs a simplified rectangular window with 8-bit fixed-point weights. Additionally, variations in update resolution and timing contribute to these differences.

%Achieving comparable learning behavior on hardware will require tuning of parameters such as learning rates, window sizes, and initial weight distributions in both the simulator and FPGA environment. This demonstrates the importance of algorithm–hardware co-design, and highlights how NeuroCoreX enables iterative evaluation of learning dynamics under realistic hardware constraints. Future work will focus on tuning the system for MicroSeer and scaling to larger datasets on more advanced neuromorphic platforms.

The trained SNN model was first developed in the SuperNeuroMAT simulator and then ported to NeuroCoreX for hardware execution. When STDP was disabled, the network outputs from NeuroCoreX closely matched those produced by the simulator, demonstrating functional equivalence in inference. However, when STDP was enabled, a divergence in weight evolution and learning behavior was observed. This discrepancy stems from two primary sources: (1) SuperNeuro uses an exponential STDP learning rule with 64-bit floating-point precision, while NeuroCoreX implements a simplified rectangular learning window with signed 8-bit fixed-point weight representation; and (2) differences in numerical resolution and synaptic update timing result in non-identical learning trajectories.
To achieve comparable accuracy metrics across simulation and hardware, tuning of learning parameters—such as learning rate, window size, and initial weight distributions—is required in both environments. These results underscore the importance of algorithm–hardware co-design in bridging the gap between simulation and deployment for neuromorphic graph learning. NeuroCoreX enables iterative testing and refinement of learning dynamics under realistic hardware constraints, facilitating the transition from simulated models to deployable systems. Future work will focus on tuning the system for MicroSeer and scaling to larger datasets on more advanced neuromorphic platforms.

The total on-chip power consumption of the NeuroCoreX design was estimated at 305~mW, with 75\% attributed to dynamic power. The Mixed-Mode Clock Manager (MMCM) accounted for the largest portion of dynamic power, followed by BRAMs, reflecting the memory-intensive nature of synaptic storage and buffering.

\section{Discussion and Conclusion}
\label{sec:conclusion}
% Other fpga hardware designs:
% NeuroCoreX demonstrates the feasibility of implementing biologically inspired spiking neural networks (SNNs) on low-cost FPGA hardware using a digital VHDL-based approach. The system supports real-time operation, online learning through spike-timing-dependent plasticity (STDP), and full all-to-all connectivity, all within a compact and open-source design. By leveraging time-multiplexing and pipelining, NeuroCoreX emulates a 100-neuron network using a single physical processing unit, making efficient use of limited hardware resources.
% This design provides an accessible platform for both neuromorphic research and hands-on educational use. It paves way for transition from simulations on SNN simulator to quick FPGA-based prototyping. In the experiments performed, network architecture and model parameters were exported directly from the SuperNeuroMAT and transferred to the FPGA via the UART interface, demonstrating a seamless transition from software-based SNN modeling to FPGA implementation. More broadly, scalable neuromorphic systems must address hardware constraints such as fixed-point representations, limited weight precision, and simplified learning rules, which often diverge from the floating-point and high-complexity models used in software simulations. NeuroCoreX supports such co-design by enabling early-stage evaluation of these constraints, making it a valuable tool for aligning algorithmic development with hardware feasibility.

NeuroCoreX enables real-time SNN emulation with STDP learning on low-cost FPGAs through a compact, open-source VHDL design. It bridges the gap between simulation and hardware by supporting seamless transfer of network models from SuperNeuroMAT, facilitating early-stage algorithm–hardware co-design under practical constraints like fixed-point precision and limited memory. Several key design choices in NeuroCoreX reflect critical trade-offs between biological realism, hardware efficiency, and scalability.

\subsection{Scalability}
One such decision is the use of an all-to-all connected network, which introduces significant memory overhead for weight storage—particularly as the network size increases. While this raises questions about scalability, the choice was made to provide maximum flexibility for users to implement arbitrary network architectures, and it is essential for supporting densely connected networks.
An alternative approach would be to store a list of addresses corresponding to either all outgoing or incoming neurons for each neuron. This is more efficient for sparsely connected networks, as it avoids storing unnecessary weights. However, it becomes highly inefficient for densely connected networks, as it requires additional BRAM to store connection addresses alongside the weights.
Since real-world networks often include a mix of dense and sparse connectivity, a scalable solution should combine both approaches. In such a hybrid model, densely connected sub-networks can be mapped directly onto NeuroCoreX’s all-to-all weight matrix, while sparse connections between these sub-networks can be implemented across multiple physical instances of NeuroCoreX. Each instance would represent a densely connected region of the larger network. With this strategy, the NeuroCoreX platform can be scaled effectively to support larger, more biologically realistic networks. We plan to implement this hybrid approach in future iterations of NeuroCoreX.

\subsection{Scalability and Network Acceleration Trade-offs}

NeuroCoreX employs time-multiplexing to emulate 100 virtual neurons using a single physical processing unit, significantly reducing hardware resource usage. However, this approach introduces a fundamental trade-off between the number of neurons that can be emulated, the achievable acceleration timescale, and the clock frequency used. Increasing the number of virtual neurons directly increases the time required to complete a single simulation step unless the update clock is proportionally scaled.

In the Artix-7 FPGA used for this work, the 100-neuron configuration utilizes nearly the full BRAM capacity due to storage requirements for a) the FIFO buffer that stores and transmits input spike trains to the network in real time, b) FIFO to store membrane potential, and synaptic current for the time multiplexed implementation, c) FIFO to transmit spike trains of all neurons back to PC, d) synaptic weight matrices 
$W_{AA}$ and $W_{in}$, e) the update and enable-STDP masks, and f) the pre- and post-synaptic trace registers required for learning. Even with more memory on higher-end FPGAs, scalability is constrained by the interaction between memory bandwidth and update clock frequency. BRAM access is sequential, with only one read or write per port per cycle, creating a bottleneck during real-time weight updates. For biologically realistic timing, the memory clock must run significantly faster than the neuron update clock. In our system, BRAM is accessed at 100~MHz while neuron updates proceed at 100~kHz, enabling a 1000\(\times\) speedup.

Assuming BRAM size is not a limiting factor, the theoretical upper bound for network size under this configuration is approximately 500 neurons. This limit arises from the need for up to 500 cycles each for STDP potentiation and depression, which cannot occur concurrently on the same matrix. Beyond this, the system cannot complete all required memory operations within a single time step unless either the clock domains are further decoupled or memory bandwidth is increased.

These constraints underscore a core design trade-off between network size, update speed, and biological fidelity. Depending on the target application, designers must balance these factors—favoring high-throughput inference, real-time learning, or alignment with biologically realistic timescales.

\subsection{Applications}

The flexibility, low cost, and on-chip learning capabilities of NeuroCoreX make it well suited for edge AI applications such as robotics, neuromorphic sensing, and adaptive control. Its energy-efficient architecture supports real-time inference and learning without reliance on offline processing. In educational settings, NeuroCoreX offers a unique platform for students to bridge the gap between software simulation and digital hardware design. By working directly with spiking neuron models and real-time hardware behavior, students gain hands-on experience with both computational neuroscience concepts and hardware engineering practices. %As an open-source tool, it promotes reproducibility, benchmarking, and community-driven development under realistic hardware constraints.
As an open-source framework, NeuroCoreX also encourages community-driven extensions, reproducibility of results, and transparent benchmarking of neuromorphic algorithms under realistic hardware constraints.

Several FPGA-based neuromorphic hardware platforms have been reported in the literature~\cite{szczerek2025quarter}, each targeting different performance and application trade-offs. Caspian~\cite{mitchell2020caspian}, for instance, has been widely used in edge applications involving control and classification tasks~\cite{patton2022neuromorphic}. It supports configurable, though limited, connectivity and can be deployed on low-cost FPGAs. Liu et al.~\cite{liu2022fpga} proposed NHAP, a platform optimized for high-speed and low-power execution of SNNs, supporting multiple neuron models on mid- to high-end FPGAs. Matinizadeh et al.~\cite{matinizadeh2024neuromorphic} introduced SONIC, an end-to-end framework for neuromorphic computing on FPGAs, which includes a core SNN engine called QUANTISENC, written in VHDL. This framework demonstrates several classification and control applications using fixed-topology layered networks.

None of the existing FPGA-based neuromorphic platforms offer the full set of features provided by NeuroCoreX—most notably, on-chip learning via STDP. While some focus on high-speed inference or limited configurability, they typically lack embedded learning and depend on offline training. Most also lack a fully open-source toolchain, limiting their adoption in research and education. In contrast, NeuroCoreX combines real-time operation, on-chip STDP, flexible connectivity, and open-source support, making it a unique platform for deployment and algorithm–hardware co-design.

%None of the existing FPGA-based neuromorphic platforms incorporate the full range of capabilities offered by NeuroCoreX—most notably, the support for on-chip learning via STDP. While many designs focus on high-speed inference or support limited connectivity, they typically lack hardware-embedded learning and rely on offline weight training. In addition, most platforms do not provide a fully open-source toolchain, limiting reproducibility and broader adoption in academic and research environments. In contrast, NeuroCoreX integrates real-time operation, on-chip STDP learning, fully configurable connectivity, and an open-source workflow, making it a distinct and versatile platform for both practical deployment and algorithm–hardware co-design.

%Despite these advances, none of the above platforms support on-chip learning through synaptic plasticity mechanisms such as STDP. Moreover, most do not provide a fully open-source toolchain, limiting accessibility and broader adoption in research and educational settings. In contrast, NeuroCoreX combines real-time operation, on-chip learning, full connectivity, and an open-source workflow—making it a unique platform for both practical deployment and algorithm–hardware co-design.

\subsection{Future Work}
Future enhancements to NeuroCoreX will focus on improving scalability and runtime efficiency. To support faster synaptic updates during STDP, we plan to explore multi-port BRAM architectures~\cite{lin2015bram}, which can alleviate memory bandwidth limitations by allowing concurrent read and write access to synaptic data. The modular design of NeuroCoreX also allows for replacing the current LIF neuron model with more biologically plausible alternatives~\cite{gautam2022conductance, gerstner1995time}, enabling broader exploration of neural dynamics in hardware.

%Future enhancements to NeuroCoreX will focus on improving scalability and runtime efficiency. %Increasing parallelism by unrolling the time-multiplexed neuron loop would allow simultaneous updates of multiple neurons, reducing emulation time at the cost of increased logic and memory usage. 

Higher-order STDP rules, such as triplet-based~\cite{gautam2024suppression} and neuromodulated STDP~\cite{fremaux2016neuromodulated}, have shown comparable learning performance to conventional STDP while requiring as little as 3-bit fixed-point resolution per synapse. Integrating these models may slightly increase logic complexity but could significantly reduce BRAM usage, enabling simulation of larger networks. Additionally, exposing pipeline depth and clock parameters as configurable settings would allow fine-tuning for task-specific trade-offs between performance and biological fidelity.

%Several enhancements could improve the scalability and performance of NeuroCoreX. One approach is to increase parallelism by partially or fully unrolling the time-multiplexed neuron loop, allowing multiple neurons to update simultaneously. While this demands more logic and memory, it can reduce emulation time for larger networks. Multi-port BRAM architectures~\cite{lin2015bram} could also ease bandwidth constraints by supporting concurrent read/write access to synaptic data.

%Support for richer plasticity models, such as triplet-based STDP~\cite{gautam2024suppression, babadi2016stability} or neuromodulated STDP~\cite{fremaux2016neuromodulated}, could be added with modest resource overhead. Finally, pipeline depth and clock rates could be made configurable, allowing the system to adapt to application-specific trade-offs between speed, learning capability, and biological fidelity.
NeuroCoreX provides an efficient, accessible, and extensible platform for implementing biologically inspired, energy-efficient SNNs on FPGA hardware. With its modular VHDL design, real-time STDP learning, and flexible connectivity, it enables researchers and educators to explore neuromorphic principles under realistic hardware constraints. Our results show that SNN execution on NeuroCoreX aligns closely with software simulations in SuperNeuroMAT, enabling seamless mapping from software to hardware. By supporting this direct transition, NeuroCoreX facilitates algorithm–hardware co-design and paves the way for scalable, low-power, and adaptive AI systems. As an open-source tool, it establishes a foundation for community-driven advances at the intersection of neuroscience, engineering, and embedded intelligence.
%NeuroCoreX provides an efficient, accessible, and extensible platform for implementing biologically inspired energy-efficient SNNs on FPGA hardware. Through its modular VHDL design, real-time STDP learning, and flexible connectivity, it enables researchers and educators to explore neuromorphic principles under realistic hardware constraints. We have also shown that NeuroCoreX's SNN execution matches the software execution in the SuperNeuroMAT simulation, thereby enabling seamless mapping from software to hardware. By supporting direct transition from software simulators to hardware execution, NeuroCoreX facilitates the co-design of algorithms and architectures, paving the way for scalable, low-power, and adaptive AI systems. As an open-source tool, it lays the foundation for broader community involvement in advancing neuromorphic computing at the intersection of neuroscience, engineering, and embedded intelligence.
%\section*{Acknowledgment}

\bibliographystyle{plain}
\bibliography{Refs}

\begin{thebibliography}{10}

\bibitem{aimone2022review}
James~B Aimone, Prasanna Date, Gabriel~A Fonseca-Guerra, Kathleen~E Hamilton,
  Kyle Henke, Bill Kay, Garrett~T Kenyon, Shruti~R Kulkarni, Susan~M
  Mniszewski, Maryam Parsa, et~al.
\newblock A review of non-cognitive applications for neuromorphic computing.
\newblock {\em Neuromorphic Computing and Engineering}, 2(3):032003, 2022.

\bibitem{optical_recognition_of_handwritten_digits_80}
E.~Alpaydin and C.~Kaynak.
\newblock {Optical Recognition of Handwritten Digits}.
\newblock UCI Machine Learning Repository, 1998.
\newblock {DOI}: https://doi.org/10.24432/C50P49.

\bibitem{benjamin2021neurogrid}
Ben~Varkey Benjamin, Nicholas~A Steinmetz, Nick~N Oza, Jose~J Aguayo, and
  Kwabena Boahen.
\newblock Neurogrid simulates cortical cell-types, active dendrites, and
  top-down attention.
\newblock {\em Neuromorphic Computing and Engineering}, 1(1):013001, 2021.

\bibitem{caragea2014citeseer}
Cornelia Caragea, Jian Wu, Alina Ciobanu, Kyle Williams, Juan
  Fern{\'a}ndez-Ram{\'\i}rez, Hung-Hsuan Chen, Zhaohui Wu, and Lee Giles.
\newblock Citeseer x: A scholarly big dataset.
\newblock In {\em Advances in Information Retrieval: 36th European Conference
  on IR Research, ECIR 2014, Amsterdam, The Netherlands, April 13-16, 2014.
  Proceedings 36}, pages 311--322. Springer, 2014.

\bibitem{cassidy2011combinational}
Andrew Cassidy, Andreas~G Andreou, and Julius Georgiou.
\newblock A combinational digital logic approach to stdp.
\newblock In {\em 2011 IEEE international Symposium of Circuits and Systems
  (ISCAS)}, pages 673--676. IEEE, 2011.

\bibitem{cong2023hyperparameter}
Guojing Cong, Shruti Kulkarni, Seung-Hwan Lim, Prasanna Date, Shay Snyder,
  Maryam Parsa, Dominic Kennedy, and Catherine Schuman.
\newblock Hyperparameter optimization and feature inclusion in graph neural
  networks for spiking implementation.
\newblock In {\em 2023 International Conference on Machine Learning and
  Applications (ICMLA)}, pages 1541--1546. IEEE, 2023.

\bibitem{cong2022semi}
Guojing Cong, Seung-Hwan Lim, Shruti Kulkarni, Prasanna Date, Thomas Potok,
  Shay Snyder, Maryam Parsa, and Catherine Schuman.
\newblock Semi-supervised graph structure learning on neuromorphic computers.
\newblock In {\em Proceedings of the International Conference on Neuromorphic
  Systems 2022}, pages 1--4, 2022.

\bibitem{date2023superneuro}
Prasanna Date, Chathika Gunaratne, Shruti R.~Kulkarni, Robert Patton, Mark
  Coletti, and Thomas Potok.
\newblock Superneuro: A fast and scalable simulator for neuromorphic computing.
\newblock In {\em Proceedings of the 2023 International Conference on
  Neuromorphic Systems}, pages 1--4, 2023.

\bibitem{date2021computational}
Prasanna Date, Bill Kay, Catherine Schuman, Robert Patton, and Thomas Potok.
\newblock Computational complexity of neuromorphic algorithms.
\newblock In {\em International Conference on Neuromorphic Systems 2021}, pages
  1--7, 2021.

\bibitem{date2023encoding}
Prasanna Date, Shruti Kulkarni, Aaron Young, Catherine Schuman, Thomas Potok,
  and Jeffrey Vetter.
\newblock Encoding integers and rationals on neuromorphic computers using
  virtual neuron.
\newblock {\em Scientific Reports}, 13(1):10975, 2023.

\bibitem{date2022virtual}
Prasanna Date, Shruti Kulkarni, Aaron Young, Catherine Schuman, Thomas Potok,
  and Jeffrey~S Vetter.
\newblock Virtual neuron: A neuromorphic approach for encoding numbers.
\newblock In {\em 2022 IEEE International Conference on Rebooting Computing
  (ICRC)}, pages 100--105. IEEE, 2022.

\bibitem{date2022neuromorphic}
Prasanna Date, Thomas Potok, Catherine Schuman, and Bill Kay.
\newblock Neuromorphic computing is turing-complete.
\newblock In {\em Proceedings of the International Conference on Neuromorphic
  Systems 2022}, pages 1--10, 2022.

\bibitem{davies2018loihi}
Mike Davies, Narayan Srinivasa, Tsung-Han Lin, Gautham Chinya, Yongqiang Cao,
  Sri~Harsha Choday, Georgios Dimou, Prasad Joshi, Nabil Imam, Shweta Jain,
  et~al.
\newblock Loihi: A neuromorphic manycore processor with on-chip learning.
\newblock {\em Ieee Micro}, 38(1):82--99, 2018.

\bibitem{fremaux2016neuromodulated}
Nicolas Fr{\'e}maux and Wulfram Gerstner.
\newblock Neuromodulated spike-timing-dependent plasticity, and theory of
  three-factor learning rules.
\newblock {\em Frontiers in neural circuits}, 9:85, 2016.

\bibitem{gautam2021adaptive}
Ashish Gautam and Takashi Kohno.
\newblock An adaptive stdp learning rule for neuromorphic systems.
\newblock {\em Frontiers in Neuroscience}, 15:741116, 2021.

\bibitem{gautam2022conductance}
Ashish Gautam and Takashi Kohno.
\newblock A conductance-based silicon synapse circuit.
\newblock {\em Biomimetics}, 7(4):246, 2022.

\bibitem{gautam2023adaptive}
Ashish Gautam and Takashi Kohno.
\newblock Adaptive stdp-based on-chip spike pattern detection.
\newblock {\em Frontiers in Neuroscience}, 17, 2023.

\bibitem{gautam2024suppression}
Ashish Gautam, Takashi Kohno, Prasanna Date, Robert Patton, and Thomas Potok.
\newblock A suppression-based stdp rule resilient to jitter noise in spike
  patterns for neuromorphic computing.
\newblock In {\em 2024 International Conference on Neuromorphic Systems
  (ICONS)}, pages 209--216. IEEE, 2024.

\bibitem{gerstner1995time}
Wulfram Gerstner.
\newblock Time structure of the activity in neural network models.
\newblock {\em Physical review E}, 51(1):738, 1995.

\bibitem{kulkarni2023sensor}
Shruti~R. Kulkarni, Aaron Young, Prasanna Date, Narasinga Rao~Miniskar, Jeffrey
  Vetter, Farah Fahim, Benjamin Parpillon, Jennet Dickinson, Nhan Tran, Jieun
  Yoo, et~al.
\newblock On-sensor data filtering using neuromorphic computing for high energy
  physics experiments.
\newblock In {\em Proceedings of the 2023 International Conference on
  Neuromorphic Systems}, pages 1--8, 2023.

\bibitem{lin2015bram}
Jiun-Liang Lin and Bo-Cheng~Charles Lai.
\newblock Bram efficient multi-ported memory on fpga.
\newblock In {\em VLSI Design, Automation and Test (VLSI-DAT)}, pages 1--4.
  IEEE, 2015.

\bibitem{liu2022fpga}
Yijun Liu, Yuehai Chen, Wujian Ye, and Yu~Gui.
\newblock Fpga-nhap: A general fpga-based neuromorphic hardware acceleration
  platform with high speed and low power.
\newblock {\em IEEE Transactions on Circuits and Systems I: Regular Papers},
  69(6):2553--2566, 2022.

\bibitem{maheshwari2023fpga}
Disha Maheshwari, Aaron Young, Prasanna Date, Shruti Kulkarni, Brett
  Witherspoon, and Narsinga~Rao Miniskar.
\newblock An fpga-based neuromorphic processor with all-to-all connectivity.
\newblock In {\em 2023 IEEE International Conference on Rebooting Computing
  (ICRC)}, pages 1--5. IEEE, 2023.

\bibitem{markram1997regulation}
Henry Markram, Joachim L{\"u}bke, Michael Frotscher, and Bert Sakmann.
\newblock Regulation of synaptic efficacy by coincidence of postsynaptic aps
  and epsps.
\newblock {\em Science}, 275(5297):213--215, 1997.

\bibitem{matinizadeh2024neuromorphic}
Shadi Matinizadeh, Arghavan Mohammadhassani, Noah Pacik-Nelson, Ioannis
  Polykretis, Krupa Tishbi, Suman Kumar, ML~Varshika, Abhishek~Kumar Mishra,
  Nagarajan Kandasamy, James Shackleford, et~al.
\newblock Neuromorphic computing for the masses.
\newblock In {\em 2024 International Conference on Neuromorphic Systems
  (ICONS)}, pages 39--46. IEEE, 2024.

\bibitem{mayr2019spinnaker}
Christian Mayr, Sebastian Hoeppner, and Steve Furber.
\newblock Spinnaker 2: A 10 million core processor system for brain simulation
  and machine learning.
\newblock {\em arXiv preprint arXiv:1911.02385}, 2019.

\bibitem{merolla2014million}
Paul~A Merolla, John~V Arthur, Rodrigo Alvarez-Icaza, Andrew~S Cassidy, Jun
  Sawada, Filipp Akopyan, Bryan~L Jackson, Nabil Imam, Chen Guo, Yutaka
  Nakamura, et~al.
\newblock A million spiking-neuron integrated circuit with a scalable
  communication network and interface.
\newblock {\em Science}, 345(6197):668--673, 2014.

\bibitem{miniskar2024neuro}
Narsinga~Rao Miniskar, Aaron~R Young, Kazi Asifuzzaman, Shruti Kulkarni,
  Prasanna Date, Alice Bean, and Jeffrey~S Vetter.
\newblock Neuro-spark: A submicrosecond spiking neural networks architecture
  for in-sensor filtering.
\newblock In {\em 2024 International Conference on Neuromorphic Systems
  (ICONS)}, pages 63--70. IEEE, 2024.

\bibitem{mitchell2020caspian}
J~Parker Mitchell, Catherine~D Schuman, Robert~M Patton, and Thomas~E Potok.
\newblock Caspian: A neuromorphic development platform.
\newblock In {\em Proceedings of the 2020 Annual Neuro-Inspired Computational
  Elements Workshop}, pages 1--6, 2020.

\bibitem{patton2022neuromorphic}
Robert Patton, Prasanna Date, Shruti Kulkarni, Chathika Gunaratne, Seung-Hwan
  Lim, Guojing Cong, Steven~R Young, Mark Coletti, Thomas~E Potok, and
  Catherine~D Schuman.
\newblock Neuromorphic computing for scientific applications.
\newblock In {\em 2022 IEEE/ACM Redefining Scalability for Diversely
  Heterogeneous Architectures Workshop (RSDHA)}, pages 22--28. IEEE, 2022.

\bibitem{schuman2021sparse}
Catherine~D Schuman, Bill Kay, Prasanna Date, Ramakrishnan Kannan, Piyush Sao,
  and Thomas~E Potok.
\newblock Sparse binary matrix-vector multiplication on neuromorphic computers.
\newblock In {\em 2021 IEEE International Parallel and Distributed Processing
  Symposium Workshops (IPDPSW)}, pages 308--311. IEEE, 2021.

\bibitem{stimberg2019brian}
Marcel Stimberg, Romain Brette, and Dan~FM Goodman.
\newblock Brian 2, an intuitive and efficient neural simulator.
\newblock {\em elife}, 8:e47314, 2019.

\bibitem{szczerek2025quarter}
Wiktor~J Szczerek and Artur Podobas.
\newblock A quarter of a century of neuromorphic architectures on fpgas--an
  overview.
\newblock {\em arXiv preprint arXiv:2502.20415}, 2025.

\bibitem{wurm2023arithmetic}
Ahna Wurm, Rebecca Seay, Prasanna Date, Shruti Kulkarni, Aaron Young, and
  Jeffrey Vetter.
\newblock Arithmetic primitives for efficient neuromorphic computing.
\newblock In {\em 2023 IEEE International Conference on Rebooting Computing
  (ICRC)}, pages 1--5. IEEE, 2023.

\end{thebibliography}

\vspace{12pt}
\color{red}

\end{document}